\documentclass[preprint,12pt,authoryear]{elsarticle}

\usepackage{amssymb}
\usepackage{amsmath}
\usepackage{algorithm}
\usepackage{algpseudocode}
\usepackage{multirow}
\usepackage{todonotes}
\usepackage[inline]{enumitem}
\usepackage[bookmarks=true]{hyperref}
\usepackage{color, colortbl}
\usepackage{subcaption,graphicx}

\journal{Robotics and Autonomous Systems}

\begin{document}

\begin{frontmatter}



\title{dGrasp: NeRF-Informed Implicit Grasp Policies with Supervised Optimization Slopes}

\author[hka,kit]{Gergely Sóti}
\author[kit]{Xi Huang}
\author[hka]{Christian Wurll}
\author[hka,kit]{Björn Hein}

\affiliation[hka]{organization={Karlsruhe University of Applied Sciences, Institute for Robotics and Autonomous Systems},
            city={Karlsruhe},
            postcode={76133}, 
            country={Germany}}

\affiliation[kit]{organization={Karlsruhe Institute of Technology, Institute of Anthropomatics and Robotics},
            city={Karlsruhe},
            postcode={76131}, 
            country={Germany}}

\begin{abstract}
We present dGrasp, an implicit grasp policy with an enhanced optimization landscape. This landscape is defined by a NeRF-informed grasp value function. The neural network representing this function is trained on simulated grasp demonstrations. During training, we use an auxiliary loss to guide not only the weight updates of this network but also the update how the slope of the optimization landscape changes. This loss is computed on the demonstrated grasp trajectory and the gradients of the landscape. With second order optimization, we incorporate valuable information from the trajectory as well as facilitate the optimization process of the implicit policy. Experiments demonstrate that employing this auxiliary loss improves policies' performance in simulation as well as their zero-shot transfer to the real-world.
\end{abstract}




\end{frontmatter}

\section{Introduction}
Robotic grasping is a fundamental task in the automation of object manipulation. Despite extensive research and progress, dealing with unknown objects under real-world conditions is still a major challenge. In this context, learning from demonstration (LfD) has recently become an attractive alternative to reinforcement learning (RL) for policy learning due to its advantages in bypassing the need for a reward function and its higher sample efficiency. LfD does not require exploration or extensive data gathering but instead learns directly from high-quality demonstration data.

\begin{figure}[tbp]
    \begin{subfigure}[b]{0.33\linewidth}        
        \centering
        \includegraphics[width=1.\linewidth]{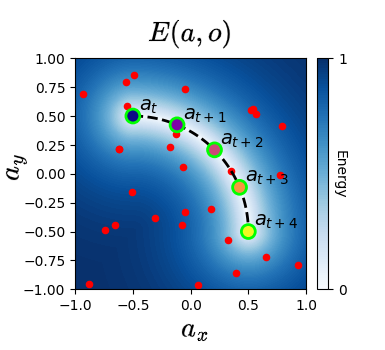}
        \caption{Implicit Behavior Cloning}
        \label{fig:ibc_policy}
    \end{subfigure}\hfill 
    \begin{subfigure}[b]{0.33\linewidth}        
        \centering
        \includegraphics[width=1.\linewidth]{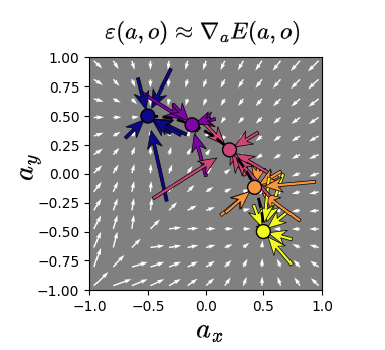}
        \caption{Diffusion Policy}
        \label{fig:diff_policy}
    \end{subfigure}\hfill 
    \begin{subfigure}[b]{0.33\linewidth}     
        \centering
        \includegraphics[width=1.\linewidth]{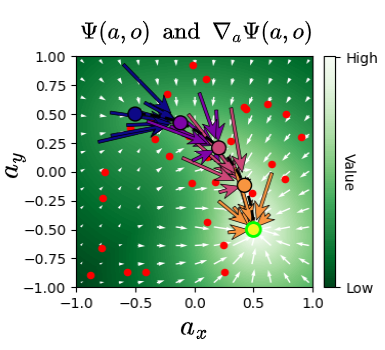}
        \caption{Our Approach}
        \label{fig:our_policy}
    \end{subfigure} 
    \caption[]{\textbf{Policy Representations} - Comparison of policy representations for observation $o$ and a demonstration trajectory $\{a_t, a_{t+1}, ... , a_{t+4}\}$ in a two-dimensional action space.
    \begin{enumerate*}[label=(\alph*)]
        \item \textit{Implicit Behavior Cloning (IBC)} learns an energy function $E(a, o)$ using negative sampling;
        \item \textit{Diffusion Policy} learns a noise function $\varepsilon(a, o)$ that approximates the gradient field of the energy function $\nabla_a E(a,o)$;
        \item In \textit{our approach} we learn a value function using negative sampling $\Psi(a, o)$ additionally supervising its gradients $\nabla_a \Psi(a, o)$ using the demonstration trajectories during training.
    \end{enumerate*} 
    This formulation combines the convenient representation of IBC and the stability and robustnes of diffusion policy.
}
    \label{fig:policy_comparison}
\end{figure}

Within Learning from Demonstration (LfD), implicit behavior cloning (IBC, \cite{florence2022implicit}) and diffusion policies (\cite{chi2023diffusion}) have emerged as effective methods. Both utilize optimization-based policies but with different underlying mechanisms. IBC learns an energy function over the joint action-observation distribution (Figure \ref{fig:ibc_policy}), which is minimized during inference to determine robot actions, effectively framing the policy as an optimization problem. This formulation allows the use of convenient action spaces like the 6-DoF task space of a robot, facilitating the incorporation of advanced scene representations such as Neural Radiance Fields (NeRFs, \cite{mildenhall2020nerf, lin2023vision}), as demonstrated by \cite{soti20246dof}. These integrations lead to improved generalization and enable zero-shot sim-to-real transfer of grasp policies. However, training IBC policies requires negative sampling for normalization, which can lead to instability, as noted by \cite{florence2022implicit} and \cite{du2020improved}.

Diffusion policies address this instability by learning a gradient field and using a denoising process to generate robot actions (Figure \ref{fig:diff_policy}). The gradient field is approximated directly by a noise prediction network, avoiding the need for normalization through negative sampling, offering a more stable alternative to IBC.

Our approach seeks to maintain the benefits of implicit policies while reducing the instability from negative sampling by also supervising the gradient field of the implicit policy during training (Figure \ref{fig:our_policy}). In action spaces such as TCP poses, the gradient of the energy function in implicit policies should naturally align with the robot's movement. By extending the IBC framework to align its gradients with demonstrated robot movements, we ensure that the gradients reflect the pose changes seen in successful demonstrations. The primary goal of this work is to show that this sort of alignment of the model to the physical world in combination with NeRF based representations leads to significantly improved convergence behavior and sim-to-real transfer of implicit policies.

We summarize our main contributions as follows:
\begin{itemize}
    \item We propose a natural way to incorporate demonstration trajectories into the training of implicit policies.
    \item We introduce a simple augmentation and training mechanism to supervise the gradients of implicit policies using the trajectories.
    \item We apply this method in the NeRF-based implicit policy by \cite{soti20246dof} and demonstrate its effectiveness on simulated and real grasping tasks.
\end{itemize}

The remainder of this paper is structured as follows: Related work is discussed in Section \ref{sec:related_work}, and a summary of the background information required for a complete description of our approach is provided in Section \ref{sec:background}. This is followed by a detailed description of our method in Section \ref{sec:method}, and the conducted experiments and results in Section \ref{sec:experiments}. The paper concludes with a brief discussion of the findings, the limitations of our approach, and possible ideas for future research in Section \ref{sec:conclusion}.

\section{Related Work}
\label{sec:related_work}
Data-driven policies in the context of robotic grasping are widely researched topic with a variety of approaches as detailed in the surveys by \cite{bohg2013data, kleeberger2020survey, newbury2023deep}. Broadly, these methods fall into four categories:
\begin{enumerate*}[label=(\roman*)]
  \item object-detection based (\cite{tobin2017domain, dong2019ppr, kleeberger2020single}),
  \item reinforcement learning (\cite{levine2018learning, song2020grasping, berscheid2021robot}), 
  \item supervised learning from a large-scale, labeled dataset (\cite{mahler2019learning, acronym2020, chen2022neural}), and
  \item learning from demonstration (\cite{zeng2021transporter, florence2022implicit, chi2023diffusion})
\end{enumerate*}.

In this work, we focus on behavior cloning, a popular end-to-end framework to learn policies from demonstrations, and even in behavior cloning there are two emerging branches, i.e. explicit and implicit models. Explicit models, like the works of \cite{avigal2022speedfolding, florence2019self, rahmatizadeh2018vision, zeng2021transporter}, propose actions directly from observations. Implicit models on the other hand learn to evaluate actions and are used in conjunction with sampling based or gradient-based optimization to find optimal actions (\cite{florence2022implicit, soti20246dof}). In this paper, we aim to extend the training process of such implicit models in a way that facilitates the optimization process, and thus improves the policy. In the following we review related work in the context of implicit policies and finally we discuss the core idea of our proposed approach. 

\cite{florence2022implicit} investigate the effects of using implicit models for behavior cloning (IBC) across a variety of robot policy learning tasks. They define an implicit policy as the $\mathrm{argmin}$ of a continuous energy function, which is learned from demonstrations. This energy function is expected to assign lower energies to optimal actions like the demonstrations, and higher energies otherwise. To find minimum locations and thus optimal actions, they propose two sampling-based algorithms and a gradient-based algorithm. Their research shows that implicit models provide competitive results or outperform explicit models and reinforcement learning algorithms on complex, discontinuous and multi-modal simulated and even real robotic tasks. 

With a similar framework, \cite{soti20246dof} learns 3 degree-of-freedom (DoF) grasps in simulation and applies zero-shot transfer to the real world. Unlike IBC, which minimizes an energy function, these methods maximize a value function and use a gradient-based optimization with the Adam optimizer \cite{kingma2014adam} instead of the gradient-based Langevin sampling described in IBC. An additional key characteristic of the approach is the usage of a pre-trained Neural Radiance Field (NeRF) \cite{mildenhall2020nerf, lin2023vision} to inform the implicit model model and thus requiring only RGB observations during inference and enabling a large degree of generalization.

NeRFs themselves learn an implicit representation of the environment, and have been used in various works involving robotic grasping (\cite{ichnowski2021dex, kerr2022evo, dai2023graspnerf, blukis2023oneshot}). However, these applications primarily leverage NeRFs for augmenting observations or as feature extractors for explicit policies, differing from the implicit optimization-based framework we use.

Although Neural Motion Fields, by \cite{chen2022neural}, do not employ learning from demonstration, they also use an implicit model to learn a grasp value function and to generate grasp trajectories in an object-centric way. In this approach, training the grasp value function requires a curated set of ground truth grasp poses and their model requires a segmented point cloud as input. Trajectories are generated by optimizing the learned implicit value function via sampling-based model predictive control.

Related to implicit models, \cite{weng2023neural} approach grasp synthesis by predicting the distance of an action candidate to the nearest successful grasp and minimize this distance to achieve successful grasps. By integrating this distance metric into CHOMP motion planning as an additional cost, the model can generate grasp trajectories. For training, this method also requires a large grasp dataset (\cite{acronym2020}) and the model processes a segmented point cloud as input.

Training the energy based model for implicit polices often involves a negative sampling process, which is konwn to cause training instability (\cite{florence2022implicit, du2020improved}). To avoid this, 
diffusion policies by \cite{chi2023diffusion}, use a denoising neural network that captures the gradient of the reversed diffusion process. They train their model by minimizing the difference of a diffused action from ground truth and a synthetic denoised action. Given a sequence of observations, the policy uses the gradient field iteratively to denoise a sequence of randomly sampled actions and finally execute them. While the diffusion policy learns the gradient field to update action sequences from noisy ones, the landscape of implicit models captures the slope from an arbitrary state towards an optimal action. This means, that the optimization process of an implicit policy itself results in an action sequence.

In this current approach, our goal is to improve the optimization landscape of the implicit grasp policy described in \cite{soti20246dof}. In addition to learning from the grasp pose itself, we use second-order optimization to supervise the gradients of the optimization landscape using the demonstrated grasp trajectory. Although, using a second-order optimization algorithm like BFGS algorithm (\cite{fletcher2000practical}) might prove beneficial for training, such algorithms usually involve the computation of the Hessian and often its inverse, too, which can be expensive. Instead, we use the Adam optimizer for both training and to lead the pose along the landscape during optimization, consistent with \cite{yen2021inerf, soti2023gradient, soti20246dof}.

\section{Background}
\label{sec:background}

We build on a previous work that introduces the concept of using transfer learning from Neural Radiance Fields (NeRFs, \cite{mildenhall2020nerf}) to train a grasp value function that can be used in an implicit policy (\cite{soti20246dof}). In this section we briefly introduce NeRFs and describe the architecture used by \cite{soti20246dof} to compute the grasp value, the way it is trained and used in an implicit policy framework to infer grasp poses.

\subsection{Neural Radiance Field - NeRF}
\label{subsec:nerf}
NeRFs, (\cite{mildenhall2020nerf}) have revolutionized scene representation by learning implicit 3D structures from 2D images. Originally developed for novel view synthesis, NeRFs combine traditional rendering methods with deep learning to achieve impressive performance. They map positions and direction vectors, representing the 5-DoF space, to color and density values, which are then used in a volume rendering pipeline to render pixels for a camera image. The key innovation of NeRFs lies in their ability to accurately synthesize new views of a scene by learning a continuous volumetric scene function from a sparse set of 2D images. This scene function acts as a powerful scene representation, capturing the detailed geometry and appearance of the environment. 

\subsection{Transfer Learning with NeRFs for Grasp Value Function}
\label{subsec:grasp-nerf}
Leveraging the scene representation provided by NeRFs, \cite{soti20246dof} introduce a method that applies transfer learning to an image-conditioned NeRF variant (\cite{lin2023vision}), using it to inform a grasp value function within an implicit policy framework. The grasp value function, denoted as $\Psi$, maps 6-DoF Tool Center Point (TCP) poses $p$ and observations $o$ to scalar values, with higher values indicating a greater likelihood of a successful grasp. The grasp policy, $\pi$, is formulated as follows:
\begin{equation}
\pi(o) = \mathrm{argmax}_p \Psi(p, o)
\end{equation}
Here, $\pi(o)$ represents an estimated optimal TCP pose for grasping. To find these maximum locations of $\Psi$, gradient-based optimization is employed.

\subsubsection{Architecture}
\begin{figure}[tbp]
    \centering
  \includegraphics[width=1.\linewidth]{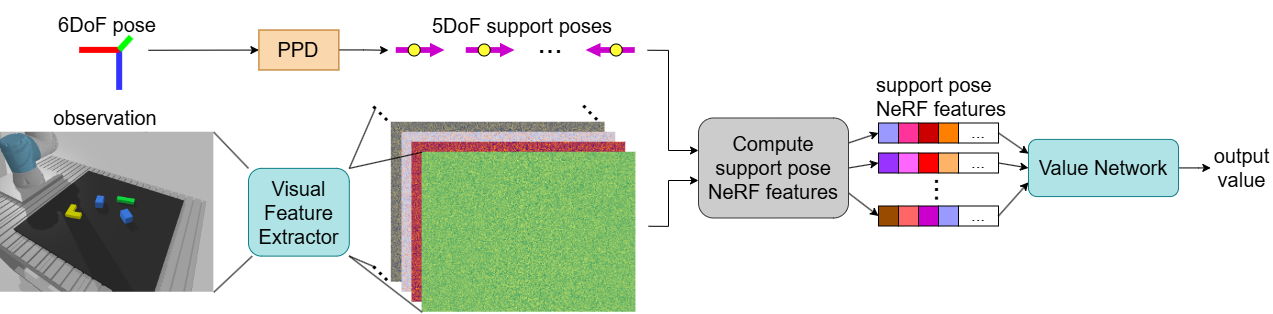}
\caption{\textbf{Computational model for the implicit policy's value function $\Psi$ for a 6-DoF grasp candidate and an observation} - First, partial pose decomposition (PPD) is applied to the 6-DoF grasp candidate to obtain a set of 5-DoF support poses, and a feature map is computed from the input observation. Then, for each 5-DoF support pose, a NeRF feature vector is computed using the extracted feature map and a pre-trained NeRF. These are finally processed by the value network to obtain the grasp value for the input 6-DoF grasp candidate.}
\label{fig:value_computation}
\end{figure}

$\Psi$ itself consists of four modules: partial pose decomposition (PPD), a visual feature extractor, a module to compute NeRF features using a pre-trained image-conditioned NeRF, and a value network. Figure \ref{fig:value_computation} illustrates their interaction to compute the value of a 6-DoF grasp candidate given an observation.

\begin{figure}[tbp]
    \centering
  \includegraphics[width=0.5\linewidth]{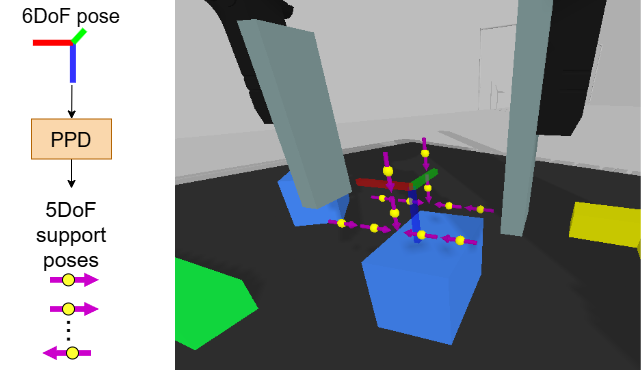}
\caption{\textbf{Partial pose decomposition} - A set of 5-DoF support poses are computed from an initial 6-DoF pose using predefined transformations. The image shows the TCP as a 6-DoF pose and a possible set of its support poses that correspond to the gripper's geometry: the yellow points with purple direction vectors pointing inwards to characterize possible object boundaries that the gripper could grasp.}
\label{fig:ppd}
\end{figure}

\textbf{Partial Pose Decomposition (PPD)} - NeRFs process 5-DoF poses to obtain color and density values for novel view synthesis. To evaluate 6-DoF grasp candidates, PPD is applied. This computes a set of predefined 5-DoF support poses from a 6-DoF pose, which can be processed independently by the NeRF and aggregated later to characterize the initial 6-DoF pose. Figure \ref{fig:ppd} shows a possible PPD for grasping that corresponds to the geometry of the gripper.

\textbf{Image-Conditioned NeRF} - An image-conditioned NeRF (VisionNeRF by \cite{lin2023vision}) is used to compute feature vectors for each support pose. In this context, image conditioning means that during training and inference, a visual feature extractor processes input observations and informs the NeRF of the current scene. This results in a NeRF that can be used in multiple environments without retraining based on a set of observations, ensuring consistent representation across different scenes.

\begin{figure}[tbp]
    \centering
  \includegraphics[width=1.\linewidth]{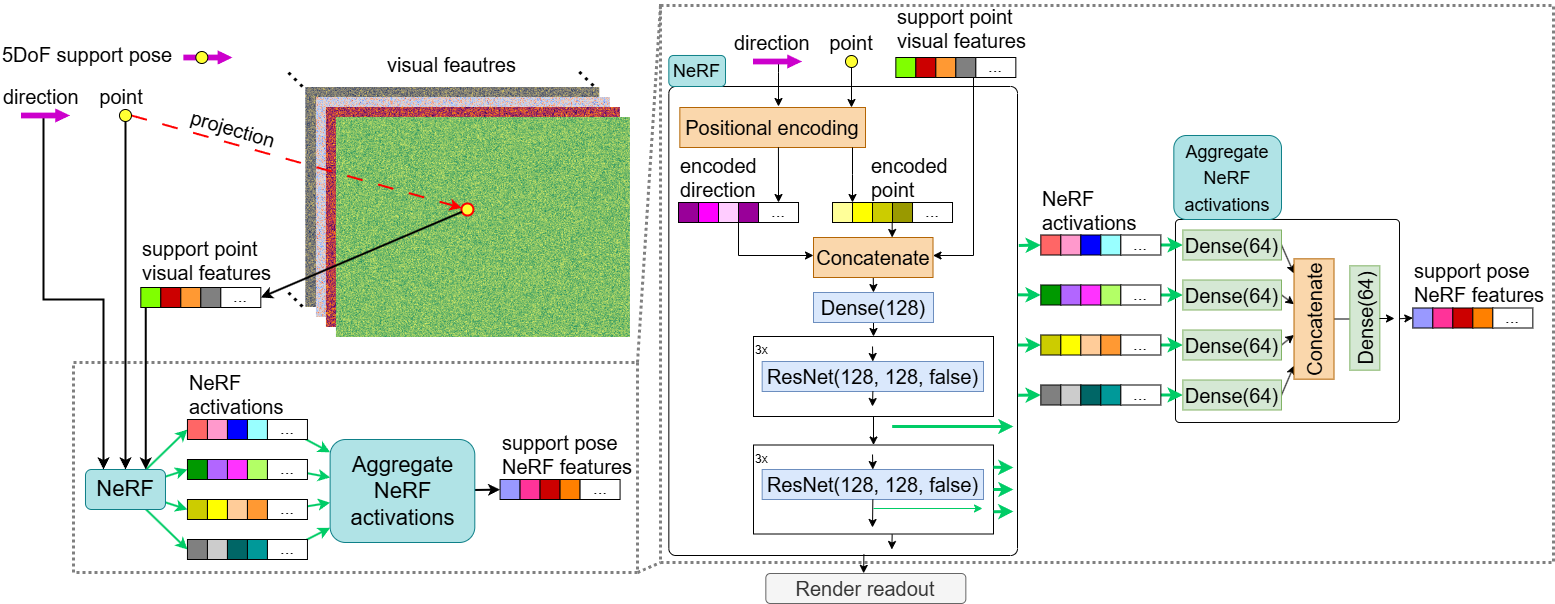}
\caption{\textbf{Computation of support pose NeRF features} - Left: Computation of NeRF features for a 5-DoF support pose using its corresponding visual feature vector; Right: network architecture of the NeRF and activation aggregation models. The green arrows represent the activations of the NeRF's last four ResNet blocks that are aggregated to form the NeRF feature corresponding to the input support pose.
}
\label{fig:support_features}
\end{figure}

\begin{figure}[tbp]
    \begin{subfigure}[b]{0.48\linewidth}        
        \centering
        \includegraphics[width=.9\linewidth]{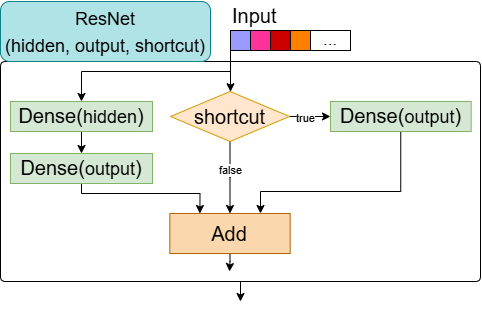}
        \caption{ResNet}
        \label{fig:resnet}
    \end{subfigure}\hfill 
    \begin{subfigure}[b]{0.48\linewidth}     
        \centering
        \includegraphics[width=.9\linewidth]{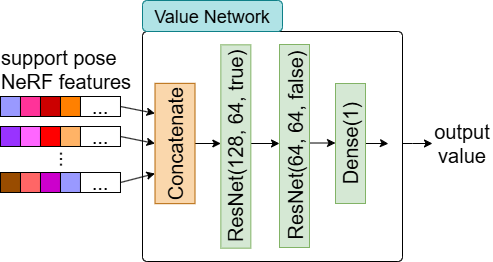}
        \caption{Value network}
        \label{fig:value_network}
    \end{subfigure} 
    \caption{\textbf{Network architectures} - Value Network: The network processes the support pose NeRF features to compute the final grasp value for the implicit policy; ResNet: Used in NeRF and Value Networks. Transforms the residual shortcut if the input and output dimensions are not equal. }
    \label{fig:architectures}
\end{figure}

\textbf{Visual Feature Extractor and Support Pose NeRF Features} - The visual feature extractor combines a pre-trained vision transformer with fully convolutional neural networks as described by \cite{lin2023vision} to compute features from input observations to inform the NeRF. The extracted features correspond to the same perspective that input was provided from. Using the camera's calibration information, the 3D point of a support pose is projected onto the extracted feature maps to obtain visual feature vectors, as shown in Figure \ref{fig:support_features}. The support poses, along with their corresponding visual feature vectors, are processed by the NeRF. A positional encoding typical for NeRFs is applied to both the 3D point and the direction vectors of the support pose:
\begin{equation}
    \gamma(v) = (sin(2^0\pi v), cos(2^0\pi v), ..., sin(2^{M-1}\pi v), cos(2^{M-1}\pi v))
\end{equation}
The function $\gamma$ is applied to each vector dimension separately, and the results are concatenated. The NeRF itself contains six ResNet blocks  (see Figure \ref{fig:resnet} for ResNet architecture) and is pre-trained with randomized scenes for novel view synthesis. To inform the value network, the NeRF's activations after the last four ResNet blocks are used. These are aggregated into the NeRF feature vector characterizing the support pose. Figure \ref{fig:support_features} shows the architecture for computing the NeRF features for a support pose given the extracted visual features from the observation.

\textbf{Value Function} - Finally, the support pose NeRF features are processed by the value function to obtain the value for the implicit policy. Figure \ref{fig:value_network} shows the network architecture representing the value function.

\subsubsection{Training}

The implicit policy's value function $\Psi$ is trained using demonstrations by transforming the learning process into a binary classification problem. Each demonstrated grasp is labeled with $1$, and randomly sampled poses within the workspace serve as negative examples labeled with $0$. The input observations $o$ are camera images of the scene with known calibration before executing the grasp. With this setup, the categorical cross-entropy loss function is used during training:
\begin{equation}
\label{eq:value_loss}
\mathcal{L}_{value} = -\log \frac{e^{\Psi(p^0, o)}}{ \sum_{i=0}^{N} e^{\Psi(p^i, o)}}
\end{equation}
Here, $p^0$ is the successful demonstration and $p^i$ (with $i \in [1,N]$) are the sampled negative examples. This way, the model is trained to assign higher scores to demonstrated poses compared to other poses within the workspace.

\subsubsection{Optimization Process}
The optimization process adapts a set of randomly sampled initial input poses to maximize the output value of $\Psi$, as outlined in Algorithm \ref{alg:grasp-opt}. A pose consists of a 3D position vector and the quaternion representation of its orientation. First, the translations are optimized for 16 iterations, then the quaternions for 16 iterations, both adapted via the Adam optimizer (\cite{kingma2014adam}). After each optimization step, the quaternions are normalized.

\begin{algorithm}
\caption{
\small Grasp pose optimization (\cite{soti20246dof})}
\begin{algorithmic}[1]
\small
\Require Observation $o$
\Ensure Successful grasp $p^*$
\State $G \gets \text{RandomGraspCandidates()}$ \Comment{Initialization}
\While{Not Terminate} \Comment{Termination criterion}
    \ForAll{$p \in G$}
        \State $p \gets p + \nabla_p \Psi(p, o)$  \Comment{Maximize $\Psi$}
        \State $p \gets \text{PostProcess}(p)$  \Comment{Fix pose}
    \EndFor
\EndWhile
\State $p^* \gets \arg\max_{p \in G} \Psi(p, o)$  \Comment{Grasp with highest value}
\State \Return $p^*$
\end{algorithmic}
\label{alg:grasp-opt}
\end{algorithm}

In the following section, we describe our approach to improve the learning of $\Psi$ to enhance the optimization results.

\section{Method}
\label{sec:method}

The approach in \cite{soti20246dof}, as detailed in Section \ref{sec:background}, focuses on using demonstrated grasp poses to learn the grasp value function $\Psi$. However, it does not utilize the demonstrated trajectories, which encompass the entire Tool Center Point (TCP) movement.

We aim to incorporate the demonstrated grasp trajectories into the learning $\Psi$ by proposing an augmented loss function that includes an auxiliary loss term. This term aligns the gradients of $\Psi$ with demonstrated TCP movements, ensuring the grasp value function considers the entire TCP trajectory. The intuition behind this is that these gradients should ideally reflect the actual TCP movements observed in successful demonstrations. We hypothesize that this augmented loss improves the grasp value function's alignment with the physical world, leading to better optimization and improved grasp outcomes.

In the following, we detail our proposed enhancements, including the formulation of the auxiliary loss, architectural modifications, and implementation details.

\subsection{Auxiliary Loss}
\label{subsec:aux_loss}
In \cite{soti20246dof} the optimization landscape for pose optimization is shaped by the value loss $\mathcal{L}_{value}$ (Equation \ref{eq:value_loss}), which only considers the executed grasp pose. To include the grasp trajectories, we augment the grasp value function $\Psi$ to have the following property:
\begin{equation}
    p_{t+1} = p_{t} + \nabla_p \Psi(p_t, o)
\end{equation}
with $p_t$ as the TCP pose at timestep $t$ and $p_{t+1}$ as the pose at a later timestep $t+1$ during a demonstration. On one hand, this aligns with the gradient-based optimization of the grasp candidates poses (see Algorithm \ref{alg:grasp-opt}). On the other hand, given that we have access to ground truth trajectories from demonstrations, this gradient can be supervised by the displacement of the TCP pose along the trajectory during training as an auxiliary loss:
\begin{equation}
\label{eq:aux_loss}
\mathcal{L}_{aux} = -S_C(p_{t+1} \ominus p_{t}, \nabla_p \Psi(p_t, o))
\end{equation}
with $S_C$ the cosine similarity and $p_{t+1} \ominus p_{t}$ representing the element-wise difference of the pose representations. We believe this straightforward operator is effective in a gradient-based context because small gradient steps allow linear changes in the representations to be sufficient. The rationale behind using cosine similarity is that we are primarily interested in the direction of the gradients rather than their magnitude and leverage the Adam optimizer's capabilities for more stable updates.

Including the auxiliary loss, the total loss for the model:
\begin{equation}
\mathcal{L}_{total} = \mathcal{L}_{value} + \mathcal{L}_{aux}
\end{equation}

\subsection{Pose Representation}
\label{subsec:pose_representation}
The optimization process relies on the Adam optimizer to adapt the input poses of $\Psi$ in such a way that its output is maximized. This gradient-based pose optimization makes the selection of pose representation crucial.

We split the pose into a position vector and an orientation representation. While \cite{soti20246dof} use quaternions for orientation, we implement the 6D orientation representation proposed by \cite{zhou2019continuity}, which is also utilized by \cite{chi2023diffusion} for diffusion policies. For both orientation representations, it is essential to ensure they remain valid after the gradient-based update. Quaternions are normalized and for 6D representations, consisting of the first two column vectors of the rotation matrix, the vectors are orthonormalized.

To compute $\mathcal{L}_{aux}$ (Equation \ref{eq:aux_loss}), we process the cosine similarity of the gradient and $p_{t+1}\ominus p_{t}$ for the position and orientation representations independently, and then sum them. Additionally, in case of the 6D orientation representation, the two column vectors are handled independently.

It is worth noting that without the auxiliary loss, we can freely interchange pose representations, as they are only used during the optimization process and not during training. However, with the auxiliary loss, the gradient used in Equation \ref{eq:aux_loss} depends on the pose representation, requiring a new model for each representation.

\subsection{Architecture}
\label{subsec:method_architecture}
Since our goal is to incorporate the gradient of $\Psi$ into the loss function and training, we need to consider the gradients of the newly added error function $\mathcal{L}_{aux}$ with respect to the trainable weights $\theta$ of $\Psi$. When training $\Psi$, a pre-trained NeRF $\Phi$ is used with frozen weights, thus only the weights of NeRF activations aggreation network (Figure \ref{fig:support_features}) and the value network (Figure \ref{fig:value_network}) are trained and belong to $\theta$.

To compute the gradients for the weight update from $\mathcal{L}_{aux}$ we use the gradients of the grasp vale model itself, $\nabla_p \Psi(p_t, o) = \frac{\partial \Psi}{\partial p}$, which also involves evaluating the NeRF $\Phi$:
\begin{equation}
    \frac{\partial \mathcal{L}_{aux}}{\partial \theta} = \frac{\partial \mathcal{L}_{aux}}{\partial \frac{\partial \Psi}{\partial p}} \frac{\partial \frac{\partial \Psi}{\partial p}}{\partial \theta} = \frac{\partial \mathcal{L}_{aux}}{\partial \frac{\partial \Psi}{\partial p}} \frac{\partial (\frac{\partial \Psi}{\partial \Phi}\frac{\partial \Phi}{\partial p})}{\partial \theta}
\end{equation}
Since $\Phi$ is independent of $\theta$ resolving the partial differential $\frac{\partial (\frac{\partial \Psi}{\partial \Phi}\frac{\partial \Phi}{\partial p})}{\partial \theta}$ results in:
\begin{equation}
    \frac{\partial (\frac{\partial \Psi}{\partial \Phi}\frac{\partial \Phi}{\partial p})}{\partial \theta} = \frac{\partial \frac{\partial \Psi}{\partial \Phi}}{\partial \theta} \frac{\partial \Phi}{\partial p} + 0
\end{equation}
This makes:
\begin{equation}
    \frac{\partial \mathcal{L}_{aux}}{\partial \theta} = \frac{\partial \mathcal{L}_{aux}}{\partial \frac{\partial \Psi}{\partial p}} \frac{\partial ^2\Psi}{\partial \Phi \partial \theta} \frac{\partial \Phi}{\partial p}
\end{equation}

The expression shows, that $\Psi$ has a mixed partial derivative with respect to its weights $\theta$ and the activations of $\Phi$. This means, that discontinuities in its first derivative should to be avoided, otherwise its second derivative could destabilize the weight update process. Since we are in the context of neural networks, we only have to make sure, that the derivatives of the employed activation functions are continuous. In order to be able to use $\mathcal{L}_{aux}$, we replace the ReLU activation functions of the NeRF activations aggregation network (Figure \ref{fig:support_features}) and the value network (Figure \ref{fig:value_network}) with ELU. As for the NeRF $\Phi$, it is sufficient to be differentiable with respect to the input TCP pose $p$. This means that we still can use the pre-trained NeRF as it is.

\subsection{Training}
\label{subsec:implementation}
For training, both the value loss $\mathcal{L}_{value}$ and the auxiliary loss $\mathcal{L}_{aux}$ are used. For $\mathcal{L}_{value}$, the demonstrated grasp pose is augmented with negative samples as described by \cite{soti20246dof}. 

We also augment the data from the grasp trajectory to compute the auxiliary loss. For a given observation $o$ and current and future TCP poses $p_{t}$ and $p_{t+1}$, we sample a set of poses $P^r$ in the proximity of $p_{t}$. The loss is then computed on $p_{t+1} \ominus p_{r}$ and $\nabla_p \Psi(\Phi(p_r, o))$ for each sampled pose $p_r \in P^r$:

\begin{equation}
\label{eq:aux_loss_augmented}
\mathcal{L}_{aux} = \sum_{p_r \in P^r} -S_C(p_{t+1} \ominus p_{r}, \nabla_p \Psi(p_r, o))
\end{equation}

\begin{figure}[tbp]
    \centering
  \includegraphics[width=.55\linewidth]{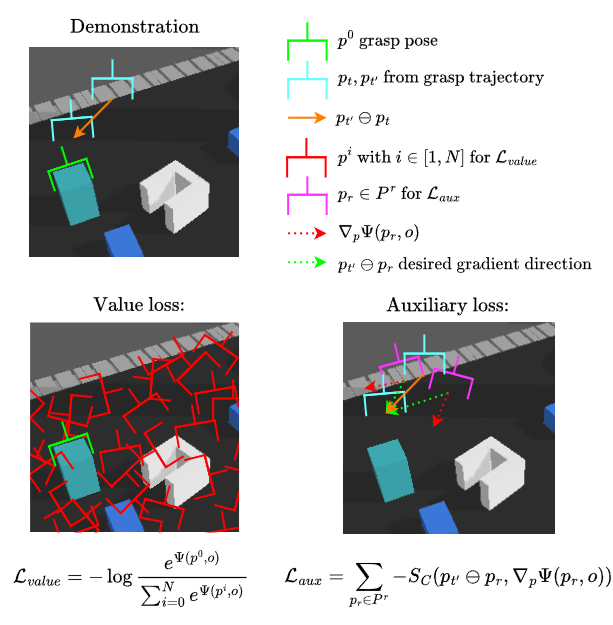}
\caption{\textbf{Loss functions} - The value loss contributes to selecting the correct grasp (green) from the randomly generated ones (red). The auxiliary loss encourages the gradients of the value function to align with the demonstrated TCP movement.}
\label{fig:losses}
\end{figure}

We rationalize this decision with when a pose is near $p_t$ then moving towards $p_{t+1}$ should still lead towards successful grasps. Figure \ref{fig:losses} visualizes the differences between $\mathcal{L}_{value}$ and $\mathcal{L}_{aux}$: $\mathcal{L}_{value}$ aims at learning to identify good grasps, and $\mathcal{L}_{aux}$ learns how to get there.

\subsection{Optimization Process}
\label{subs:pose_optimization}
The improved optimization landscape resulting from applying the auxiliary loss $\mathcal{L}_{aux}$ allows us to implement a synchronous optimization process. Instead of optimizing positions and rotations sequentially like \cite{soti20246dof}, we can now optimize both simultaneously. 

To tune the optimization process we employ Bayesian hyperparameter optimization over 100 iterations. Hyperparameters include the optimizer's initial learning rates for the pose representations, their corresponding decay rates, and the number of optimization steps.

\color{black}
\section{Experiments and Results}
\label{sec:experiments}

To evaluate the grasp policies, we conduct a series of experiments in both simulated and real-world environments. These tasks are designed to test the models' ability to generalize across familiar (in-distribution) and unfamiliar (out-of-distribution) scenarios and their adaptability to real-world conditions.
\hyphenation{dis-tri-bu-ti-on}
\hyphenation{work-space}
\subsection{Tasks}
\begin{figure}[tbp]
\begin{subfigure}[b]{0.24\linewidth}        
    \centering
  \includegraphics[width=1.0\linewidth]{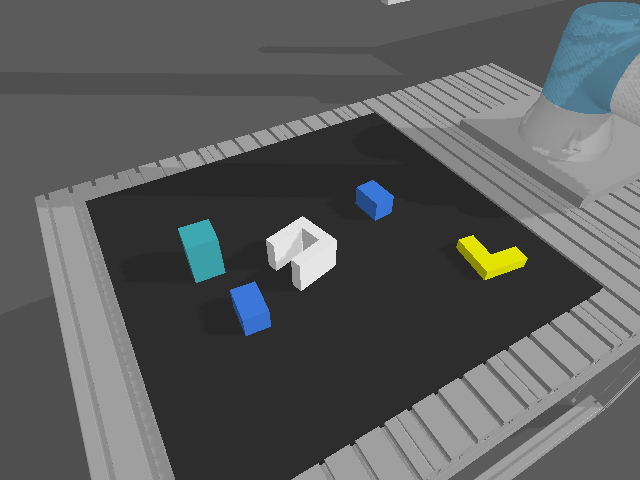}
  \caption{simple}
  \label{fig:simple_task}
\end{subfigure}\hfill 
\begin{subfigure}[b]{0.24\linewidth}     
    \centering
  \includegraphics[width=1.0\linewidth]{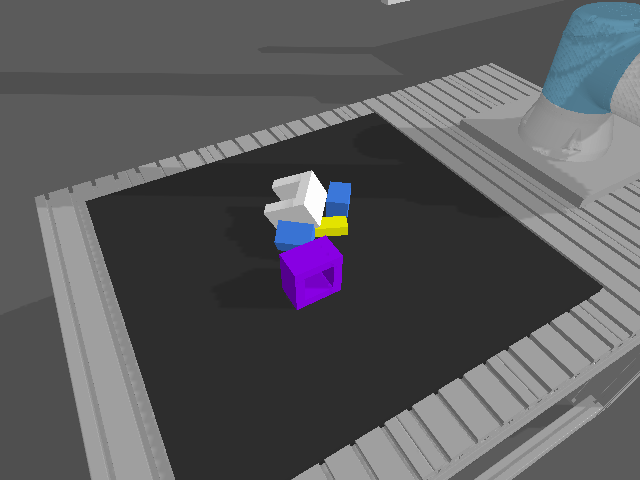}
  \caption{clutter}
  \label{fig:clutter_task}
\end{subfigure} 
\begin{subfigure}[b]{0.24\linewidth}   
    \centering
  \includegraphics[width=1.0\linewidth]{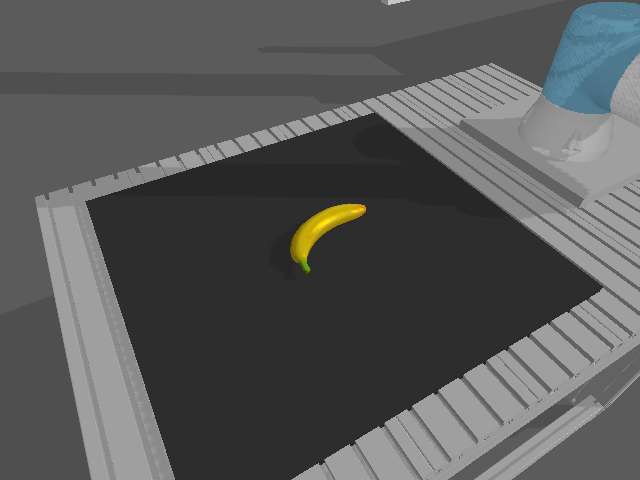}
  \caption{novel objects}
  \label{fig:ycb_task}
\end{subfigure}\hfill 
\begin{subfigure}[b]{0.24\linewidth}       
    \centering
  \includegraphics[width=1.0\linewidth]{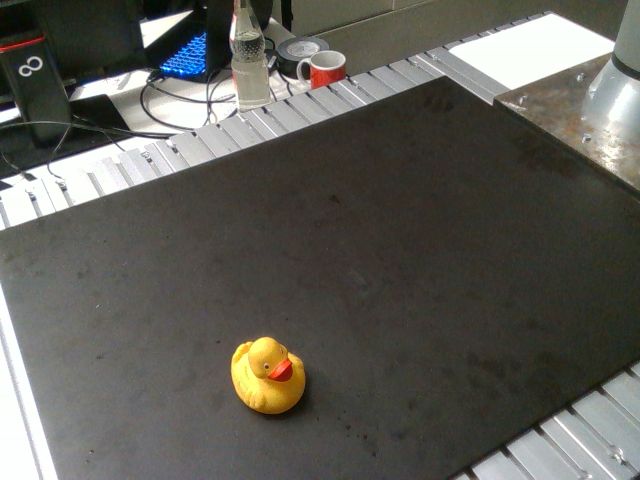}
  \caption{real-world}
  \label{fig:real_task}
\end{subfigure}

\caption{Example observations from the test datasets.}
\label{fig:tasks}
\end{figure}

We use the three simulated and a real-world task described in \cite{soti20246dof} to measure the grasp success rate of a policy for testing:
\begin{itemize}
    \item Tasks in a pybullet simulated environment. Grasping is successful if an object is enveloped by the gripper fingers and was lifted up  after the physics-based grasp execution 
    \begin{itemize}
        \item \textbf{simple:} This task assesses basic grasping capabilities. The workspace contains up to five monochromatic prismatic objects, each placed at a distance from the others (Figure \ref{fig:simple_task}). The goal is to successfully grasp one of these objects.
        \item \textbf{clutter:} This task tests the model's performance on out-of-distri-bution pose data. The scenario is a cluttered workspace containing five monochromatic prismatic objects (Figure \ref{fig:clutter_task}). The objective is to grasp all objects one after the other.
        \item \textbf{novel objects:} This task assesses the model's ability to handle out-of-distribution pose, shape, and texture data. The workspace features one previously unseen object selected from the YCB dataset \cite{calli2015ycb} (Figure \ref{fig:ycb_task}). The goal is to grasp this object. Objects used: banana, foam brick, gelatin box, hammer, Master Chef can, pear, power drill, strawberry and tennis ball.
    \end{itemize}
    \item \textbf{real-world:} This task tests the transferability of the model to the real world. In this task, a single everyday object is randomly placed in the workspace of an real robot (Figure \ref{fig:real_task}). The task is considered successful if the robot can securely grasp and lift the object. Objects used: crochet ball, a shuffled Rubik's cube, large Lego tire, canned tomato, rubber duck, hiking boot, dental floss, power drill, shampoo bottle and a 3D printed block.
\end{itemize}

All tasks feature a UR10 robot on a workbench, equipped with a Robotiq 2f-140 gripper and an Intel RealSense D415 camera. Examples of the tasks are shown in Figure \ref{fig:tasks}.

\subsection{Training and Hyperparameter Tuning}
\label{subsec:training}
During training, all models are exclusively exposed to the training dataset of the simulated \textbf{simple} scenario, including the pre-training of the NeRF and the grasp value models. This means, that neither of the models have ever seen objects in close proximity to each other, with complex textures and shapes or in poses other than upright during training.

In each of the grasp value models we use the same pre-trained NeRF, which was trained on 2.5k randomly set up \textbf{simple} tasks with 50 randomly sampled camera perspectives for 1600 epochs as described in \cite{soti20246dof}. The training data for the grasp value models consists of 512 demonstrated grasps, generated in simulation, containing RGB observations and the grasp trajectory. Successful grasp poses are determined by an oracle with access to the simulation's state. The observations are recorded from 16 randomly sampled perspectives pointing to the center of the robot's workspace before the action execution. The grasp models were trained for 400 epochs. We configure both the NeRF and grasp value models to process a single image from a single perspective at the time. 

\begin{figure}[tbp]
\begin{subfigure}[b]{\linewidth}        
    \centering
  \includegraphics[width=\linewidth]{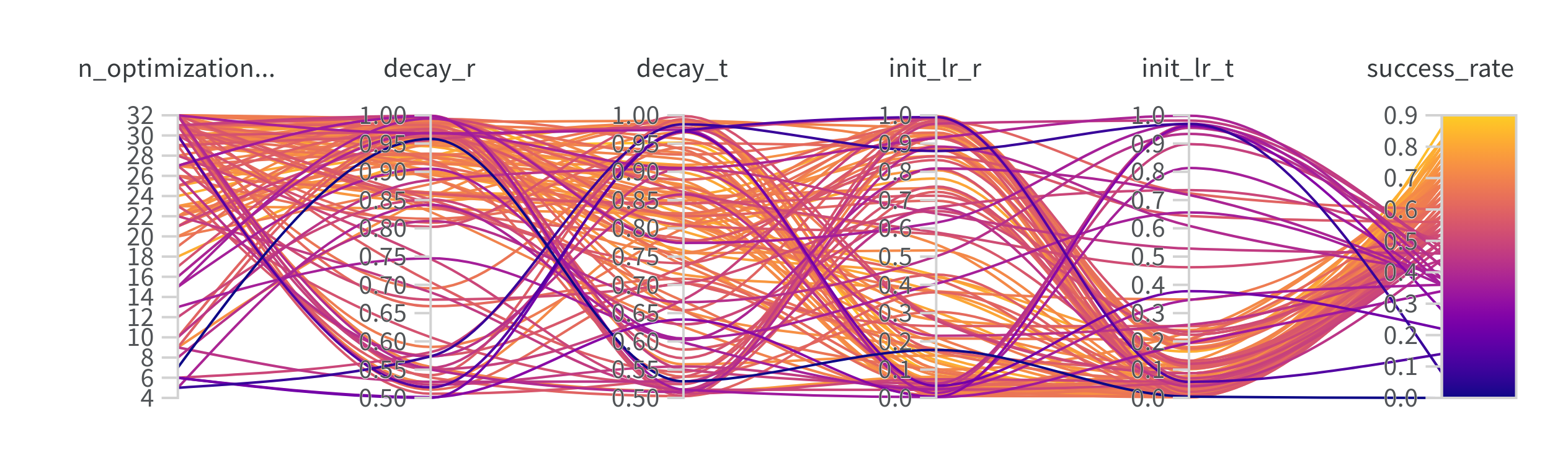}
\end{subfigure}\hfill 

\begin{subfigure}[b]{\linewidth}     
    \centering
  \includegraphics[width=\linewidth]{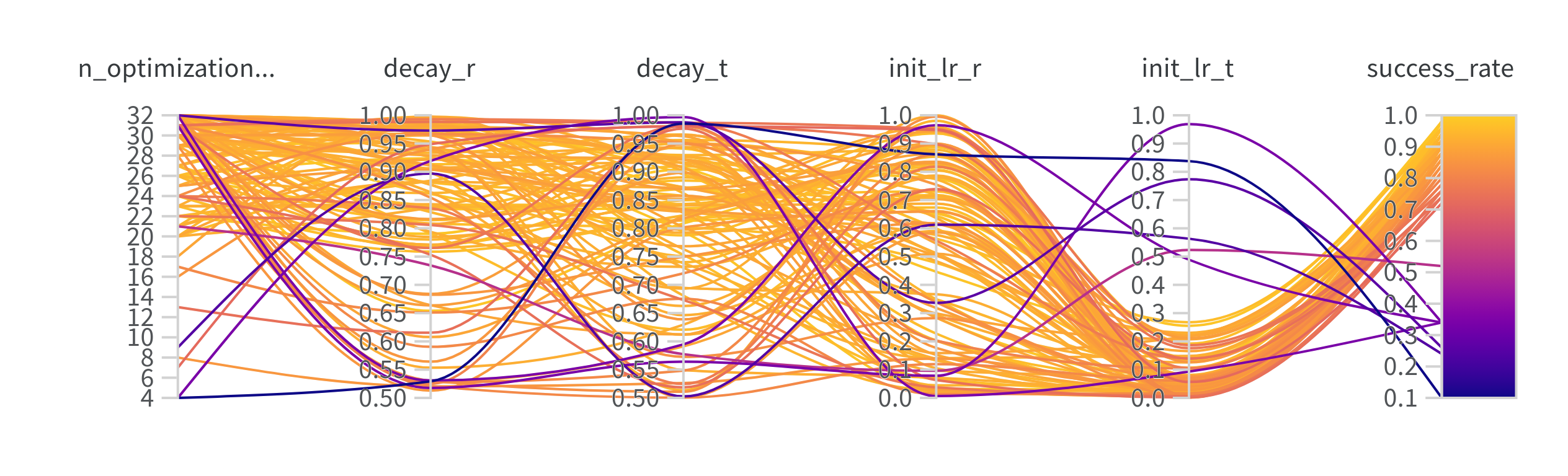}
\end{subfigure} 

\caption{\textbf{Bayesian hyperparameter optimization} - The image shows the configurations tested during hyperparameter optimization of a policy using a model trained only with the value loss $\mathcal{L}_{value}$ (top) and for a policy also employing the auxiliary loss $\mathcal{L}_{aux}$ (bottom).}
\label{fig:hyperparameter}
\end{figure}

The grasp value models can be used in combination with differently configured policies using sequential or synchronous position and orientation optimization or in case of using the value loss $\mathcal{L}_{value}$ only, the rotation representation can also be altered. After training the grasp value models we run Bayesian hyperparameter optimization to tune each policy using them. This narrows down the required number of pose optimization steps and the learning and decay rates for the pose optimization in order to improve success rates. Success rates are determined on a validation dataset of the \textbf{simple} task with simulated grasp execution. The hyperparameters tuned are described in \ref{subs:pose_optimization}. Figure  \ref{fig:hyperparameter} shows the configurations tested during tuning policies with and without employing the auxiliary loss $\mathcal{L}_{aux}$. The overall higher success rates when using $\mathcal{L}_{aux}$ indicate that the learned grasp value models are more suitable for optimization, more robust for optimization parameter perturbations and even lead to higher success rates.

\subsection{Results}
\label{sec:results}

Due to the different policy and model configurations (using $\mathcal{L}_{aux}$, rotation representation, sequential or synchronous pose optimization) and policy parameters, we trained, tuned and tested several policy and model combinations.

During our experiments we found that when not using $\mathcal{L}_{aux}$, sequential optimization in the policy leads to higher success rates, and that 6D rotation representation tends to outperform quaternions which aligns with the findings of \cite{zhou2019continuity}. When using $\mathcal{L}_{aux}$ however, quaternions perform better.

Our baseline policy \texttt{Base\textsubscript{quat}} does not use $\mathcal{L}_{aux}$ and uses quaternions as rotation representation with sequential optimization, aligning with the description in \cite{soti20246dof}. Additionally, we include \texttt{Base\textsubscript{6d}} as a second baseline, with employing 6D rotation representation as the only difference. 

In the following we present results of the best performing policy using $\mathcal{L}_{aux}$ with sequential optimization \texttt{dGrasp} and also with synchronous optimization \texttt{dGrasp\textsubscript{sync}}. Both use quaternions as rotation representation. For all models during policy inference, we use 3 input images that are processed independently by the value function but their sum is used as an objective function during optimization, as described in \cite{soti20246dof}.

\begin{table}

\begin{center}
\caption{
\label{table:all_results} \textbf{Grasp success rates} - Mean and standard deviation for simulated tasks and success rate for the real task}
\begin{tabular}{l|||c|c|c||c}
 & simple & clutter & ycb & real \\
\hline
\hline
\texttt{{Base}} & 0.77 $\pm$ 0.04 & 0.66 $\pm$ 0.03 & 0.62 $\pm$ 0.04 & 0.28 \\
\texttt{{Base\textsubscript{6d}}} & 0.79 $\pm$ 0.02 & 0.68 $\pm$ 0.03 & \textbf{0.64} $\pm$ 0.03 & 0.30 \\
\texttt{{dGrasp}} & \textbf{0.91} $\pm$ 0.02 & \textbf{0.70} $\pm$ 0.04 & 0.62 $\pm$ 0.04 & 0.56 \\
\texttt{{dGrasp\textsubscript{sync}}} & \textbf{0.91} $\pm$ 0.02 & 0.67 $\pm$ 0.01 & 0.60 $\pm$ 0.02 & \textbf{0.60} \\
\end{tabular}
\end{center}
\end{table}

\begin{figure}[tbp]
    \centering
  \includegraphics[width=.8\linewidth]{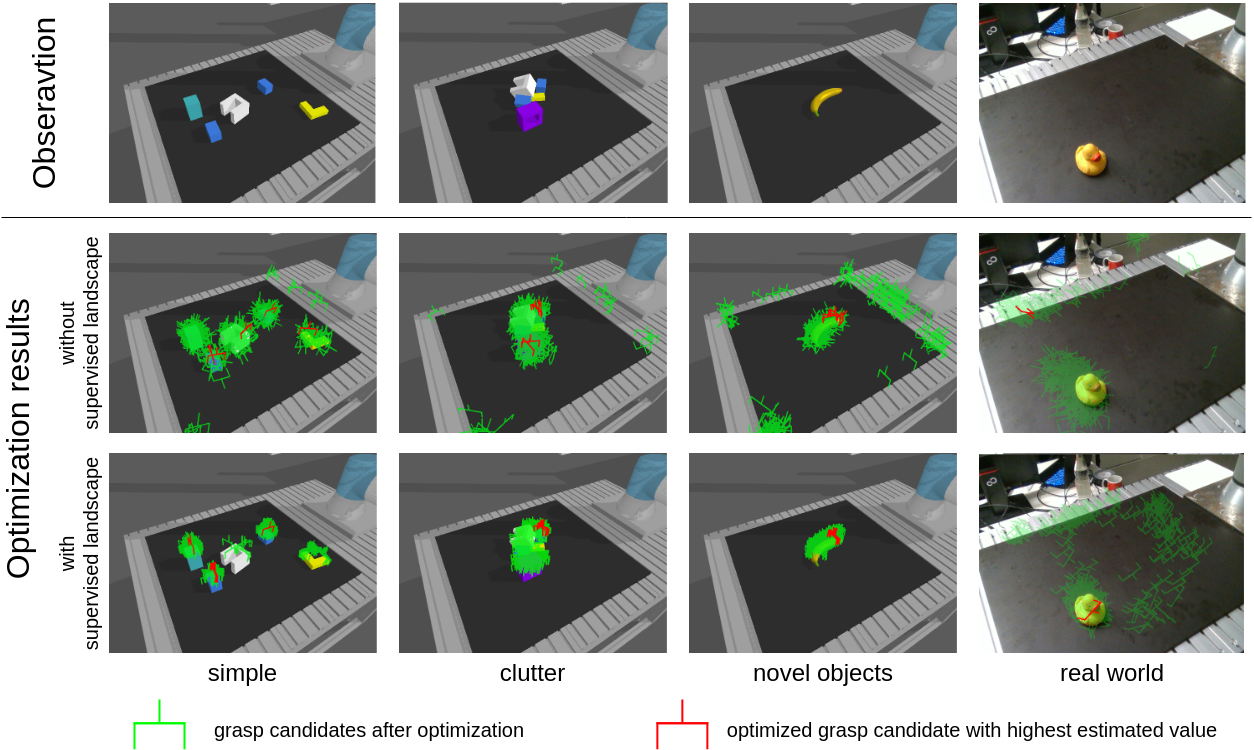}
\caption{
\textbf{Grasp pose optimization results} - The figure shows the final state of the optimization process on different tasks. The policy employing a grasp value model with a supervised optimization landscape is shown in the bottom row. The best predicted grasp candidates are highlighted in red.}
\label{fig:opt_results}
\end{figure}

The simple and ycb test datasets contain 100 different scenes where the robot is allowed to execute a single grasp. The clutter test dataset contains 20 different scenes with 5 objects in a clutter and the robot is allowed to attempt to remove an object 10 times. Grasp success is determined by execution in simulation and we repeat each simulated experiment 6 times. In the real task, each of the 10 objects is placed 5 times randomly in the robots workspace with the robot executing a single grasp every time. Table \ref{table:all_results} presents the average grasp success rates and their standard deviation in the simulated task and the success rate in the real task for the policies. An example of the final results of a policy's pose optimization for models without and with $\mathcal{L}_{aux}$ in Figure \ref{fig:opt_results} shows the improved convergence properties when supervising the optimization landscape during training, especially in simulated environments.  

There is no significant difference between model performance on the clutter and ycb tasks, however an analysis of variance (ANOVA) on the success rates yielded significant variation among the models ($F(3, 20) = 41.03, p < .001$) in case of the simple task. A post hoc Tukey test revealed that both the \texttt{trajectory-quat} model and the \texttt{trajectory-quat-sync} model had significantly higher success rates compared to the \texttt{goal-quat} and the \texttt{goal-6d} models.  This shows the effectiveness of the auxiliary loss in shaping the grasp value function while also enabling synchronous pose optimization.

Both \texttt{dGrasp} and \texttt{dGrasp\textsubscript{sync}} significantly outperform the base policies on the real task doubling the success rates. Even though we filter the final optimization results to only contain poses that point downwards (which is more of a safety feature) and adjust the final pose along the $z$ axis by 1cm to compensate the errors in camera calibration, the results indicate a significant improvement in zero-shot sim-to-real transfer capability.

Regarding the objects in the real world, the simpler objects like the rubber duck, the crochet ball and the 3D printed block were grasped reliably by all policies. All struggled however, with more complex objects like the power drill and the hiking boot. The predicted grasps commonly just collided with the objects. The policies also often failed to grasp the shampoo bottle, however mainly due to slippage. While \texttt{{Base}} and \texttt{{Base\textsubscript{6d}}} never grasped the Lego tire, the shuffled Rubik's cube, the dental floss and the canned tomato, both \texttt{dGrasp} and \texttt{dGrasp\textsubscript{sync}} were successful in 75\% of the trials. In case of the Lego tire, the \texttt{{Base}} policies ended up in one of the corners of the robots workspace, which we believe can be accounted for the tire having a similar dark color as the plate in the workspace. Both, the dental floss and the canned tomato requires precise positioning, the first due to being small and the second because it is heavy and slippery. The performance of the \texttt{dGrasp} policies on these objects indicate an improved utilization of the geometric representation provided by the NeRF.

Overall, the results show that integrating the demonstration trajectories via the auxiliary loss $\mathcal{L}_{aux}$ into the training process significantly enhances the performance of the grasping policies, particularly in zero-shot sim-to-real transfer.

\section{Conclusion}
\label{sec:conclusion}

In this work, we propose an auxiliary loss for augmenting the training of a NeRF-informed grasp value function, aimed at improving the optimization landscape of an implicit grasp policy. This auxiliary loss supervises the gradients of the value function using demonstrated grasp trajectories, and requires second order optimization during the training of the neural network model. Our experiments focus on the generalization capabilities of these policies, training models on simple simulated grasps and testing them on cluttered and novel objects in simulation and also in real-world settings through zero-shot sim-to-real transfer. The results demonstrate significant improvements in one out of three simulated tasks and in zero-shot sim-to-real transfer, suggesting that the auxiliary loss contributes to more stable policy optimization, thereby enabling more reliable identification of successful grasp poses.

Despite these advancements, the model's generalization to complex objects remains a challenge, which could likely be addressed with a more diverse dataset. While current success rates may not be sufficient for high-demand real-world applications, the promising results achieved with a small, simple dataset underscore the potential of our method.

Moreover, the proposed approach is not limited to grasping tasks and could be extended to a wide range of policies. The use of NeRFs as a scene representation, in particular, seems to significantly enhance generalization, offering the possibility of developing highly capable general models that can be easily fine-tuned for specific tasks. Although our implementation was in an open-loop context, the findings on synchronous pose optimization suggest that transitioning to a closed-loop framework could be viable, opening the door to integrating these methods into more complex and sophisticated tasks. This highlights the potential of NeRFs and our approach as a robust framework for developing more reliable, adaptable, and scalable robotic systems.

\section*{Acknowledgement}
This research is being conducted as part of the KI5GRob project funded by the German Federal Ministry of Education and Research (BMBF) under project number 13FH579KX9.

\section*{Declaration of Generative AI and AI-assisted technologies in the writing process}
Statement: During the preparation of this work the author(s) used ChatGPT in order to improve readability of the manuscript. After using this tool/service, the author(s) reviewed and edited the content as needed and take(s) full responsibility for the content of the publication.

\bibliographystyle{elsarticle-harv} 
\bibliography{references}



\end{document}